\documentclass[sigconf,nonacm]{acmart}
\usepackage{amsmath}
\usepackage{booktabs}
\usepackage{tabularx}
\usepackage{microtype}
\setcopyright{none}
\acmDOI{}
\acmISBN{}
\renewcommand\footnotetextcopyrightpermission[1]{}

\newcommand{\mse}{\operatorname{MSE}}
\newcommand{\Var}{\operatorname{Var}}
\newcommand{\Cov}{\operatorname{Cov}}
\newcommand{\Cm}{C^{m}}
\newcommand{\Sm}{S^{m}}
\newcommand{\Ch}{C^{h}}
\newcommand{\Sh}{S^{h}}

\title[What Does a Sharing Question Add?]{What Does a Sharing Question Add? Auditing LLM Survey Scores for Misinformation}

\author{Zonghuan Xu}
\affiliation{\institution{Institute of Trustworthy Embodied AI, Fudan University}
\institution{Shanghai Key Laboratory of Multimodal Embodied AI}
\city{Shanghai}\country{China}}
\author{Xiang Zheng}
\affiliation{\institution{City University of Hong Kong}\city{Hong Kong SAR}\country{China}}
\author{Yutao Wu}
\affiliation{\institution{Deakin University}\country{Australia}}
\author{Xingjun Ma}
\authornote{Corresponding author.}
\email{xingjunma@fudan.edu.cn}
\affiliation{\institution{Institute of Trustworthy Embodied AI, Fudan University}
\institution{Shanghai Key Laboratory of Multimodal Embodied AI}
\city{Shanghai}\country{China}}

\begin{document}
\begin{abstract}
Evaluating misinformation requires distinguishing whether readers believe content from whether they would share it. Asking large language models (LLMs) both questions yields two scores, but does the sharing answer contribute information beyond the credibility answer? We audit eight model versions on 290 synthetic misinformation articles, using 1,256 paired survey responses with 317 participant identifiers as an external validity criterion. An initial reversal motivates the audit: every model's raw sharing score predicts mean human sharing less accurately than its credibility score. This ordering changes after offset correction, so it does not by itself diagnose missing information. We instead distinguish score reconstructability, persistence across elicitation formats, and incremental human validity. Credibility predicts 30.8--72.5\% of model-sharing variation relative to a held-out constant baseline; remaining sharing differences correlate at 0.61--0.75 across question-order and separate-question conditions. Yet adding model sharing to human and model credibility yields only $-0.25$\% to $+0.69$\% error reduction with fixed regression, with all exploratory intervals crossing zero. Flexible prediction and format changes do not establish an improvement. Sharing answers therefore contain structured variation beyond the observed credibility score, without established incremental validity for human sharing in these data. The findings motivate validating the contribution of each elicited outcome, beyond inspecting score differences or agreement across prompts.
\end{abstract}

\maketitle

\section{Introduction}
Misinformation on the Web can be believable without being something a reader wants to share. Conversely, sharing can serve purposes other than endorsing a claim. Human studies show that directing attention toward accuracy changes sharing decisions~\cite{pennycook2021}, and self-reported sharing intentions have been associated with actual online sharing in particular settings~\cite{mosleh2020}. Evaluating responses to misleading content therefore calls for more than a single judgment of credibility. This creates a concrete measurement question as LLMs are increasingly used to answer human survey instruments: does asking the additional sharing question provide an additional useful measurement?

Using LLMs as respondents or evaluators has become a practical research strategy, with applications ranging from simulated survey samples~\cite{argyle2023} to evaluations intended to approximate users' judgments~\cite{barile2025}. Recent work directly simulates misinformation belief and sharing from participant profiles~\cite{choi2026}. These applications differ in their inputs and objectives, but all require evidence linking model answers to their intended interpretation. A model can produce two plausible survey answers without providing two distinguishable sources of human-relevant information. For Web research, this matters when model-derived measures are used to compare misleading content, pilot a survey, or inform which materials deserve further human evaluation. We investigate this question through an empirical audit of paired LLM survey scores for misinformation.

Our starting observation is counterintuitive. For the same articles, evaluator, and seven-point response scale, directly using the model's sharing answer to predict the mean human sharing response gives higher error than using its credibility answer, in all eight model versions. However, raw numerical error combines score offsets with predictive variation. Correcting only the offset reverses this ordering. The initial finding consequently raises a question about what the answers contain; it does not establish that an indirect question accesses a better model of human sharing.

We examine that question in three stages. First, can the sharing score be reconstructed from the same model's credibility score on unseen content? Second, do the remaining sharing differences persist when the question order changes or the questions are asked separately? Third, do these differences improve prediction of human sharing beyond the available credibility measurements? These are distinct requirements. Nonidentical scores need not encode distinct psychological constructs, and a reproducible scoring pattern need not align with human responses.

The data contain 290 synthetic misinformation articles from 99 claim scenarios, eight model versions, and 1,256 paired human responses linked to 317 anonymous participant codes. The models answered as ordinary readers without individual profiles. We study those answers as article-dependent measurements, using individual human records as an external criterion. The principal validity comparison asks whether model sharing adds to a predictor that already includes the same person's credibility response and model credibility. Holding out scenarios and excluding test participants from training limits reuse of content and respondent-specific information. A companion comparison excludes human credibility and evaluates the increment available from model scores alone.

We find incomplete reconstruction and partly persistent residual variation, alongside small and uncertain human-sharing increments. Separating the questions can reduce their coupling without establishing improved human validity. Our contribution is an outcome-specific audit connecting these findings: \emph{additional, partly persistent variation in LLM answers does not by itself establish an additional valid measure of human sharing}. We provide a reproducible decomposition into reconstruction, persistence, and external increment, with explicit interpretation boundaries. The findings do not identify an internal mechanism or prove zero useful information.

\section{Related Work}
\paragraph{LLMs as survey respondents.}
Argyle et al.~\cite{argyle2023} investigate simulated subpopulations by conditioning language models on demographic backgrounds. Santurkar et al.~\cite{santurkar2023} examine whose opinions model responses represent. Bisbee et al.~\cite{bisbee2024} distinguish apparent similarity in aggregate survey responses from fidelity to relationships among variables and document sensitivity to prompting and time. These studies motivate evaluating the intended use of synthetic responses rather than treating fluent answers as a sufficient criterion. Our unit of model elicitation is an article, without a specified respondent profile. Human records provide a criterion for evaluating the article scores, not matched identities that the models were asked to reproduce.

\paragraph{Survey format and psychometric validity.}
Question wording, response labels, and item order affect LLM answers~\cite{dominguez2024,tjuatja2024}. Such sensitivity creates both a concern and an opportunity for measurement: changing a format can alter the dependence between two model responses, but the change still requires external validation. A recent psychometric audit by Lukauskas and \v{S}arkauskait\.{e}~\cite{lukauskas2026} examines distributional, structural, and downstream validity of profile-conditioned synthetic survey respondents. Our narrower question concerns the contribution of an additional outcome item for the same Web content. We connect its reconstructability and cross-format residual structure to its incremental relation with human responses, rather than proposing a general psychometric score or treating model consensus as validation.

\paragraph{Belief and sharing of misinformation.}
Human credibility and sharing judgments are related but do not serve identical functions~\cite{pennycook2021}. Evidence linking intentions with actual sharing~\cite{mosleh2020} supports studying intentions while leaving their validity dependent on the instrument and context. Closest to our application, Choi et al.~\cite{choi2026} simulate misinformation belief and sharing using real respondent profiles, reporting exaggerated belief--sharing associations and distorted effects of individual and network characteristics. We do not claim that excessive coupling is a newly discovered problem. We ask whether a sharing response contains variation beyond an observed credibility response, whether that variation persists, and whether it supplies an additional externally validated signal. These questions remain relevant even when the two model outputs are visibly different.

\paragraph{Evaluating model-derived measurements.}
LLM judges are commonly assessed against human preferences~\cite{zheng2023}. LP-DIXIT checks a model-based explanation evaluation against user judgments before using it to compare explanation methods~\cite{barile2025}. Beyond Utility evaluates LLM recommenders along dimensions that ordinary accuracy metrics leave unresolved~\cite{jiang2025}. Our audit follows the same need to connect evaluation criteria to a use, while addressing a different object: paired survey scores. Reconstruction concerns dependence between model outputs; persistence concerns their behavior across elicitation formats; incremental validity concerns an external human outcome. None is a substitute for the others.

\section{Materials and Elicitation}
\subsection{Articles and Human Responses}
The stimuli originated from false or misleading claims documented by Reuters Fact Check across five topics: environment, health, innovation, livelihood, and safety. Twenty claims per topic were paired with communication goals. Each claim--goal pair defines a \emph{scenario}. Gemini 3 Pro Preview, Qwen Plus, and Qwen3-32B generated article variants under a shared template, using a single turn, temperature zero, and no external retrieval. The template framed generation as a human-judged writing competition, requested approximately 500 English words, and supplied credibility, sharing, and belief-change scoring criteria. These are purpose-built experimental texts, rather than a representative sample of naturally circulating misinformation.

After refusals, safety screening, and coverage checks, the retained set contains 290 articles from 99 scenarios, with two or three generator variants per scenario. The shared claim and goal create dependencies between variants, so scenario is the content grouping unit throughout our held-out analyses. The generator identity is not supplied as a feature in the primary audit.

Participants were recruited on Prolific in two waves. The task comprised up to five approximately 500-word articles, with an advertised duration of 15--20 minutes. The consent page warned that some texts might contain inaccurate information. Readers answered from first impressions without external lookup. Credibility asked whether the main claims felt believable and realistic; sharing asked whether readers would personally want to forward or share the text. Both used seven anchored categories. We measure stated willingness, not observed sharing.

The two items belonged to a longer instrument. Understanding preceded credibility; an attention check followed credibility; sharing preceded intent strength, belief change, and topic familiarity. Credibility thus preceded sharing for every human respondent. The model received the two focal items and anchors but not the full instrument. Any external validity established here would be conditional on these questionnaire conditions.

Participant identifiers derive from self-chosen anonymous codes; their one-to-one correspondence with platform accounts cannot be verified from the available export. We retain the first valid record per allocation and average valid allocations for each participant--article pair. The resulting 1,256 records represent 317 identifiers: 144 records in wave 1 and 1,112 in wave 2. Article coverage has median four and range one to 19. The principal analysis uses individual records, weighted equally across participant identifiers. Table~\ref{tab:scope} summarizes the data scopes; Appendix~\ref{app:elicitation} describes allocation records and Appendix~\ref{app:human-instrument} reproduces the instrument.

Retained records span March 23--28, 2026. The existing cleaning rules require a reported age of 18--80, a nonempty country field, no reported display issue, and the instructed attention-check response. After within-allocation deduplication, 1,257 allocations yield the 1,256 unique pairs. Using each participant's first retained demographic entries, the median reported age is 30 (range 18--77); the largest normalized country groups are South Africa (81), Egypt (45), and the United Kingdom (43). These entries are unverified self-reports. Thirteen participant identifiers have inconsistent ages across retained allocations, including seven with inconsistent normalized countries. We preserve the frozen primary sample and report a post-hoc exclusion sensitivity in Section~\ref{sec:robustness}. Demographic fields are not model inputs or regression covariates.

\begin{table}[t]
\caption{Audit scope. Model scores vary by article, not by human participant. Each listed model condition retains one successful response per article and item.}
\label{tab:scope}
\begin{tabularx}{\columnwidth}{@{}lX@{}}
\toprule
Component & Retained observations \\
\midrule
Content & 290 articles, 99 claim--goal scenarios \\
Human criterion & 1,256 paired records, 317 study IDs \\
Main model scores & 8 versions $\times$ 290 score pairs \\
Format comparisons & Claude: 290 articles; GPT-5.4 and Gemini 3.1 Pro: 99 each \\
Formats & $C$ then $S$; $S$ then $C$; separate calls \\
Response scale & Credibility $C$ and sharing $S$: 1--7 \\
\bottomrule
\end{tabularx}
\end{table}

\subsection{Reader-Style Model Answers}
The eight main evaluators are Claude Sonnet 4.5, Gemini 3 Pro Preview, GPT-4.1, GPT-4o, GPT-5, GPT-5.1, GPT-5.2, and GPT-5.4. The main scores were collected on March 19--25, 2026, using OpenRouter for Claude and Gemini and the OpenAI API for GPT versions. Six evaluators share a provider family; their results are not eight independent replications of a model population.

The prompt introduced an online reading survey, requested a reader's first impression, and asked for credibility followed by sharing. It required two integer fields in JSON, without an explanation. There was no system message, retrieval, or tool use. Temperature was zero for Claude, Gemini, GPT-4.1, and GPT-4o; GPT-5-family endpoints used default settings because custom temperature was not accepted. Appendix~\ref{app:elicitation} supplies exact question wording and model identifiers.

The prompt did not specify a real participant, demographic profile, social audience, or target population, and did not request a population mean. This determines the scope of the audit. We ask what information these reader-style answers provide as measurements. We do not evaluate instruction compliance with a mean-forecasting task, or fidelity to a particular human identity. In particular, human credibility used in our diagnostic regressions was never provided to the LLM.

\subsection{Existing Elicitation Variants}
Three existing conditions ask credibility then sharing, sharing then credibility, or each question in a separate call. The latter condition pairs the separate answers for the same article. Claude covers all 290 articles. GPT-5.4 and Gemini 3.1 Pro Preview cover 99 articles, one per scenario and 33 per generator; these articles have 410 human records from 232 participants. We compare formats only within their common model and article set.

These calls were collected in July 2026. Model versions, dates, or routes can differ from the main March collection, so main-to-variant comparisons do not isolate a format effect. Even within the format study, each condition has only one retained output per article and item. We can describe cross-format correspondence but cannot estimate same-condition test--retest reliability or fully separate format effects from sampling and service variation.

\section{Evaluation Design}
Let $i$ index articles, $p$ participants, and $m$ model versions. Model answers are $\Cm_i$ and $\Sm_i$; a person's paired responses are $\Ch_{pi}$ and $\Sh_{pi}$. The wave indicator is $W_{pi}$. We treat the anchored response scale as approximately interval-valued when using squared loss. The term \emph{information} below refers to the performance of specified prediction rules and the structure of observed scores, not Shannon mutual information or information in hidden representations.

\subsection{Initial Diagnostic: Raw Error and Calibration}
For the motivating comparison only, we average human sharing within each article to obtain $\bar S^h_i$ and compare raw $\Cm_i$ and $\Sm_i$ as predictions of that mean. Each article receives equal weight. A constant trained on the human means of training articles provides a feasible reference. Offset correction fits $q_i+b$; affine calibration fits $a+bq_i$, where $q$ is either model score. Ten-fold scenario-grouped validation fits these mappings without using test labels.

Raw error mixes several components. For any score $q$, the empirical identity
\begin{equation}
\begin{split}
\mse(q,\bar S^h)={}&(\bar q-\overline{\bar S^h})^2+\Var(q)\\
&+\Var(\bar S^h)-2\Cov(q,\bar S^h)
\end{split}
\label{eq:raw}
\end{equation}
shows why a large error need not imply a lack of predictive ordering. Variances and covariance here use the same article-equal empirical normalization. Calibration probes the numerical interpretation of scores; the following analyses probe what additional variation they provide. We keep these estimands separate rather than interpreting the raw reversal as the central validity test.

\subsection{Reconstructing Sharing from Credibility}
For each evaluator, we fit a map $g_m$ from $\Cm$ to $\Sm$. Outer validation leaves out entire scenarios in ten folds. A nested five-fold scenario split selects among a constant, linear and quadratic mappings, and a seven-level piecewise-linear basis with shrinkage. All selection uses training articles only. We also retain an unpenalized affine fit as a simpler reference. Predictions are clipped to the scale endpoints.

Let $\widehat S_i$ be the outer prediction of $\Sm_i$, and let $\widehat S_{0i}$ be the training-mean prediction. We report
\begin{equation}
R^2_{\mathrm{rec}}=1-\frac{\sum_i(\Sm_i-\widehat S_i)^2}
{\sum_i(\Sm_i-\widehat S_{0i})^2},
\label{eq:reconstruction}
\end{equation}
along with reconstruction RMSE on the original scale. This is a reduction relative to an out-of-fold constant, rather than an in-sample coefficient of determination. We additionally describe the within-category variation of $\Sm$ at a fixed observed $\Cm$. Failure to reconstruct a score establishes a limitation of the available score and fitted map, not an independent psychological construct.

\subsection{Persistence of Residual Score Differences}
For elicitation format $a$, subtract the mean sharing score among articles with the same credibility category:
\begin{equation}
e^a_i=S^{m,a}_i-\overline{S}^{m,a}_{C^{m,a}_i}.
\label{eq:residual}
\end{equation}
We correlate these residuals between the original order and each alternative format on the same articles. This asks whether the articles rated unusually high or low for sharing at a given credibility level retain that tendency. It removes a format-specific average relationship between the two observed scores.

Two checks use alternative controls: residuals from the held-out reconstruction maps, and residuals after projecting both sharing scores onto both formats' credibility-category bases and their numerical interaction. Cross-model comparisons apply the same controls to pairs of evaluators. These are descriptive comparisons of fixed residuals; we do not interpret a positive correlation as proof of a sharing-specific latent factor. Residual analyses were added after examining the primary results and are exploratory follow-ups.

\subsection{Incremental Validity against Human Sharing}
The primary comparison conditions on the person's observed credibility as well as model credibility. Its baseline and expanded predictors have inputs
\begin{align}
f_0 &: (\Ch_{pi},\Cm_i,W_{pi}),\nonumber\\
f_1 &: (\Ch_{pi},\Cm_i,\Sm_i,W_{pi}).
\label{eq:validity}
\end{align}
Both predict the same person's $\Sh_{pi}$. Conditioning on $\Ch$ addresses a specific criterion: does the additional model answer relate to human sharing beyond what is already available in that person's credibility judgment? This is a conditional validity diagnostic, not a deployment strategy that obtains human credibility for free. It also does not isolate a causal effect of credibility, or eliminate measurement error in that observed item.

We report a companion comparison using only $(\Cm_i,W_{pi})$ versus $(\Cm_i,\Sm_i,W_{pi})$. It tests whether the extra model item helps without observing human credibility. Model scores are constant across all readers of an article in either comparison. Evaluating individual records therefore does not turn these scores into personalized predictions.

For participant-equal weights $w_{pi}=1/n_p$, where $n_p$ is that person's number of evaluated records, define held-out risk and incremental gain as
\begin{align}
\widehat R(f)&=\frac{\sum_{p,i}w_{pi}(\Sh_{pi}-\widehat f_{pi})^2}
{\sum_{p,i}w_{pi}},\nonumber\\
\Delta&=100\frac{\widehat R(f_0)-\widehat R(f_1)}{\widehat R(f_0)}.
\label{eq:gain}
\end{align}
A positive $\Delta$ indicates improvement from adding model sharing. Its denominator is the corresponding baseline risk, not the irreducible error or a human ceiling.

\paragraph{Prediction rules.}
The fixed analysis uses weighted least squares with an intercept, wave, an unrestricted seven-level basis for human credibility, and the model scores as linear terms. A flexible analysis selects model-score terms from no model scores, linear terms, quadratic terms and interactions, and additive seven-level bases; ridge penalties are $0,1,10,100$. Human credibility and wave remain unpenalized in every candidate. The expanded candidate set contains the entire baseline set, allowing it to ignore sharing. Both procedures clip predictions to $[1,7]$. Appendix~\ref{app:estimators} specifies the bases, penalties, and tie rule.

\paragraph{Content and participant holdout.}
The main outer split groups records into ten folds by scenario. From each training fold, we additionally remove all records belonging to any test participant. Inner five-fold selection repeats both restrictions. Consequently, neither another variant of the test claim nor another response by a test participant can train the predictor or select its hyperparameters. Training weights are recomputed from each person's remaining training records. Main outer folds retain 739--794 records from 207--215 people and evaluate 122--126 records. All eligible records receive one outer prediction.

\subsection{Uncertainty and Scope of Inference}
Human validity intervals use 5,000 crossed resamples: people and scenarios are resampled independently. Each record receives the product of their multiplicities, applied to its participant-equal loss weight. The same draws compare both predictors. This accounts for the two observed clustering dimensions when resampling the fixed held-out predictions. Reconstruction and residual-correlation intervals resample scenarios. We report the 2.5th and 97.5th percentiles.

These are exploratory intervals conditional on fitted predictions or residuals. They do not refit the full selection pipeline, provide simultaneous coverage across models, or represent uncertainty over model families. A confidence interval crossing zero does not establish equivalence or absence of information. The principal protocol was specified before this analysis run but followed earlier examination of the dataset; it was not preregistered. We report every evaluator and planned sensitivity condition, rather than choosing models with favorable gains. Appendix~\ref{app:sensitivity} covers alternative weighting, wave restriction, participant exclusion, and reverse-outcome checks.

\section{Results}
\subsection{Raw Reversal Does Not Diagnose Information}
\begin{figure*}[t]
\centering
\includegraphics[width=\textwidth]{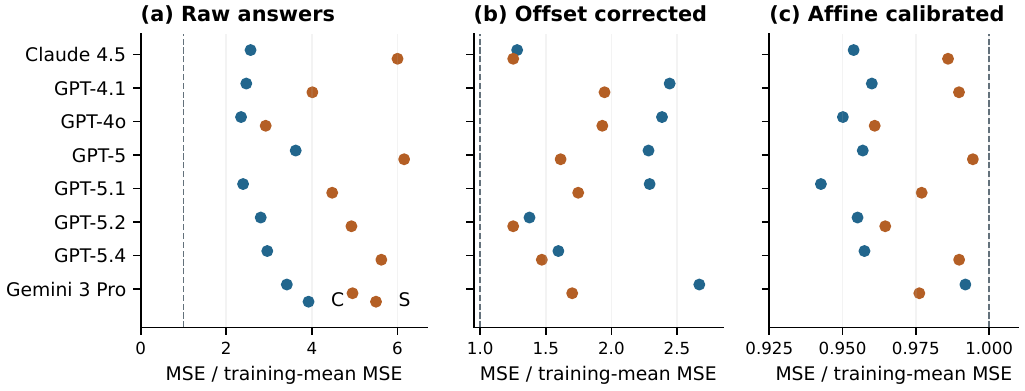}
\caption{Motivating article-mean diagnostic. Lower is better; the dashed line is the training-mean baseline. Raw $S$ is worse than $C$ for all eight evaluators, whereas correcting the offset reverses that ordering. Affine calibration brings both near the constant. Each mapping is fitted with scenario-held-out validation. Axes have different ranges to show the successive changes. This diagnostic evaluates numerical use of scores, not the conditional human-validity estimand in Equation~\ref{eq:validity}.}
\Description{Three panels compare credibility and sharing error before calibration, after offset correction, and after affine calibration across eight model versions.}
\label{fig:calibration}
\end{figure*}

All eight raw sharing scores incur higher MSE than their corresponding credibility scores when predicting article-mean human sharing (Figure~\ref{fig:calibration}a). Raw sharing MSE is 2.92--6.16 times that of the training-mean baseline. These errors are large, but the scores also have substantial numerical offsets: the scale value is not a calibrated estimate of the human mean.

Correcting only the offset changes the ordering: calibrated $S$ now has lower point-estimate error than offset-corrected $C$ for every model (Figure~\ref{fig:calibration}b). Affine calibration removes 67.1--83.9\% of raw sharing error. However, replacing the score by the training mean alone obtains 98.0--99.9\% of that error reduction. Affine $C$ again has the lower point estimate in seven evaluators, with both scores now close to the constant (Figure~\ref{fig:calibration}c). The ordering depends on how the numbers are interpreted; a large calibration improvement is not necessarily a large improvement in predicting differences between articles.

The original reversal is not unique to model answers. For 289 articles with at least two readers, we can use one randomly selected person's answer to predict the other readers' mean sharing. Averaging the squared errors over eligible reader choices gives MSE 4.772 for that person's credibility and 5.833 for their sharing. This is a comparison of one respondent with other respondents, not an optimal human forecast. It shows why neither semantic matching nor a raw $C$--$S$ reversal alone diagnoses a model-specific failure. We therefore turn to the additional variation in the model answers themselves.

\subsection{Sharing Is Not Fully Reconstructed from Credibility}
\begin{figure*}[t]
\centering
\includegraphics[width=\textwidth]{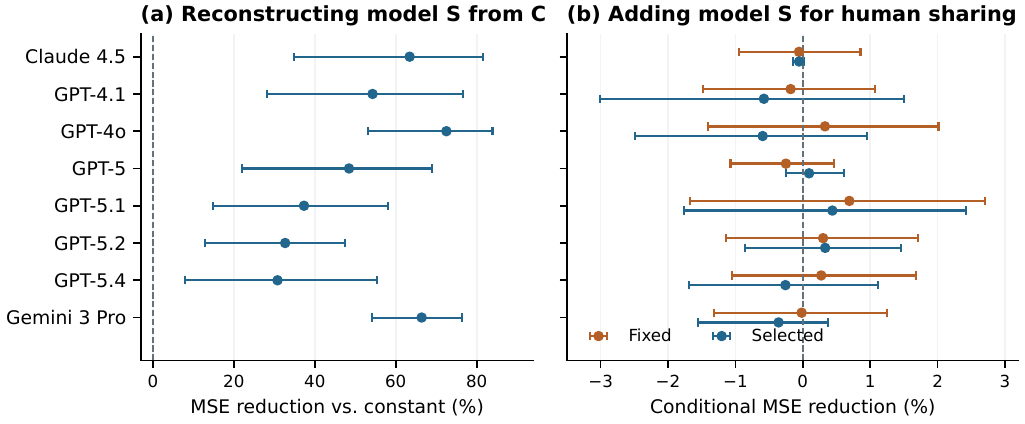}
\caption{Reconstructability and external validity answer different questions. (a) Held-out reconstruction of a model's sharing score from its own credibility score, relative to a training constant. (b) Increment from adding model sharing when predicting individual human sharing, conditional on human credibility, model credibility, and wave. Fixed and training-selected fits use scenario holdout with test-participant exclusion. Bars are exploratory 95\% intervals: scenario resampling in (a), crossed participant--scenario resampling in (b). The two panels have different targets and denominators.}
\Description{Credibility reconstructs some but not all model sharing variation. The human-validity gains cluster near zero and all intervals cross zero.}
\label{fig:information}
\end{figure*}

Credibility predicts a substantial but incomplete part of model sharing on unseen scenarios (Figure~\ref{fig:information}a). With nested selection, $R^2_{\mathrm{rec}}$ ranges from 30.8\% for GPT-5.4 to 72.5\% for GPT-4o. Reconstruction RMSE ranges from 0.439 to 0.898 scale points. For example, Claude's sharing answers are more tightly concentrated and leave a smaller residual RMSE than GPT-5.1's, even though both retain variation beyond credibility. Reconstruction percentages should therefore be read alongside residual magnitudes; neither alone measures human validity.

The incompleteness is also visible without fitting a prediction rule. Within a fixed observed credibility category, sharing still varies; the within-category variance is 25.0--63.2\% of total sharing variance across models. This descriptive calculation uses all articles, whereas Equation~\ref{eq:reconstruction} tests prediction on held-out scenarios. Together, they rule out the simplest account in which the observed seven-level credibility score uniquely determines the sharing answer on these materials.

They do not establish a separate sharing construct. Both questions could reflect a common continuous evaluation that is coarsened into seven categories. The sharing answer could reveal finer distinctions within a credibility category, or respond to additional article attributes. Variation could also include sampling noise. These possibilities motivate testing whether residual differences persist across formats before assigning them an external interpretation.

\subsection{Residual Differences Persist across Formats}
\begin{figure}[t]
\centering
\includegraphics[width=\columnwidth]{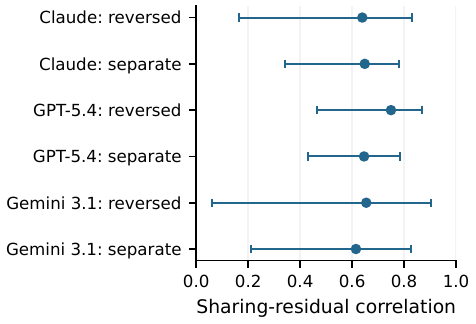}
\caption{Cross-format correspondence of sharing residuals after each condition's observed credibility categories are removed. Each comparison uses the original $C$-then-$S$ condition as the reference. Claude has 290 articles; the other models have 99. Intervals resample scenarios while holding residuals fixed. This is not same-condition test--retest reliability.}
\Description{Sharing residuals correlate positively across reversed order and separate-question conditions in three model versions.}
\label{fig:residuals}
\end{figure}

After subtracting each format's mean sharing at the observed credibility level, residual correlations across question-order and separate-question conditions range from 0.615 to 0.750 (Figure~\ref{fig:residuals}). Claude yields correlations of 0.639 for reversed order and 0.649 for separate calls. The corresponding values are 0.750 and 0.646 for GPT-5.4, and 0.654 and 0.615 for Gemini 3.1 Pro. All six exploratory scenario intervals have positive lower endpoints, although several are wide.

Using held-out reconstruction residuals gives a similar range, 0.61--0.79. Simultaneously controlling both conditions' credibility categories and their numerical interaction gives partial correlations of 0.67--0.81. The correspondence therefore survives the tested ways of removing the observed relationship with credibility. It is inconsistent with an account in which all remaining differences are independent, condition-specific noise.

Related residual structure also appears across evaluators. After controlling both models' credibility scores, the median correlation over 28 model pairs is 0.592. Excluding within-GPT-family pairs leaves 13 cross-family pairs with correlations from 0.263 to 0.630. These pairs share articles and model endpoints, so they are not independent replications. The result indicates common scoring structure, which could reflect shared article cues, coarsened common judgments, or shared bias. Human criterion validation is still needed to determine whether the structure is useful for the sharing outcome.

\begin{table*}[t]
\caption{Format-specific results on matched article sets. Reconstruction is the percentage reduction in model-sharing MSE relative to a constant, using nested selection. Human increment is the fixed-fit relative reduction from adding model sharing to human and model credibility and wave. Brackets give exploratory 95\% crossed-resampling intervals. Smaller reconstruction is not itself better: it must be interpreted alongside external validity.}
\label{tab:formats}
\centering
\begin{tabular}{llrrr}
\toprule
Evaluator & Elicitation format & Articles & Reconstruction (\%) & Human increment (\%) \\
\midrule
Claude 4.5 & $C$ then $S$ & 290 & 55.8 & -0.02 [-0.70, +0.69] \\
Claude 4.5 & $S$ then $C$ & 290 & 53.0 & +0.18 [-0.65, +1.10] \\
Claude 4.5 & Separate & 290 & 24.0 & +0.12 [-0.71, +0.99] \\
GPT-5.4 & $C$ then $S$ & 99 & 28.0 & +0.17 [-2.12, +2.05] \\
GPT-5.4 & $S$ then $C$ & 99 & 42.0 & +0.36 [-2.44, +3.05] \\
GPT-5.4 & Separate & 99 & 27.2 & -1.22 [-4.64, +0.83] \\
Gemini 3.1 Pro & $C$ then $S$ & 99 & 59.3 & -0.19 [-2.44, +1.56] \\
Gemini 3.1 Pro & $S$ then $C$ & 99 & 60.7 & -0.38 [-1.22, +0.44] \\
Gemini 3.1 Pro & Separate & 99 & 26.4 & -0.05 [-0.55, +0.39] \\
\bottomrule
\end{tabular}
\end{table*}

\subsection{Human Sharing Increment Remains Unestablished}
Figure~\ref{fig:information}b tests that external criterion. Adding model sharing to human credibility, model credibility, and wave gives relative MSE changes between $-0.25$\% and $+0.69$\% with the fixed regression. The flexible analysis gives $-0.60$\% to $+0.44$\%. Every exploratory interval crosses zero. GPT-5.1 has the largest positive main point estimate, but its interval is $[-1.67,+2.71]$\% with the fixed fit and $[-1.76,+2.43]$\% with selection. The data do not establish a positive increment for that model or a uniformly harmful increment across models.

The companion comparison excludes human credibility. Using only model scores and wave, adding sharing gives $-0.20$\% to $+0.03$\% with fixed regression and $-0.04$\% to $+0.26$\% with selection, again with all intervals crossing zero. Thus, the absence of an established gain in the main diagnostic is not solely a consequence of supplying the person's credibility response. The two comparisons nevertheless answer different questions and have different baseline errors; their percentages should not be pooled. Full per-model results appear in Appendix~\ref{app:tables}.

Allowing the expanded predictor to ignore sharing is important for interpretation. It prevents the analysis from forcing an extra score into every fit. It does not guarantee lower outer-test error: a candidate can win inner validation and generalize less well. Conversely, an exactly zero selected increment can occur when both candidate sets choose the same predictor. Neither event proves that the underlying score has exactly zero information. Our conclusion concerns the small estimated increments and their uncertainty under the tested rules.

The contrast with the preceding results is the substantive finding. The model sharing answer contains variation beyond the observed credibility category, and some of that variation persists when elicitation changes. These internal properties do not establish additional correspondence with human sharing. The result is compatible with a weak useful signal, an unmodeled conditional relationship, shared model biases, or relevant personal context missing from the prompt. The current design does not select one of these mechanisms.

\subsection{Changing Elicitation Does Not Establish a Repair}
Separate calls make sharing less reconstructable from credibility in two of the three format comparisons (Table~\ref{tab:formats}). Claude's reconstruction reduction falls from 55.8\% in the $C$-then-$S$ condition to 24.0\% under separate calls; Gemini 3.1 Pro falls from 59.3\% to 26.4\%. GPT-5.4 changes little, from 28.0\% to 27.2\%. Across models, 16--36\% of sharing answers change between the joint and separate conditions. The model outputs can therefore become less coupled without automatically becoming more useful measurements.

The associated human increments remain small or negative, with every interval crossing zero. All paired intervals for the change in increment relative to $C$ then $S$ also cross zero. Flexible fits lead to the same assessment. For example, the separate-question condition gives fixed-fit increments of $+0.12$\% for Claude, $-1.22$\% for GPT-5.4, and $-0.05$\% for Gemini 3.1 Pro. The different article counts and model versions preclude treating these values as a ranking of format effectiveness across providers.

This comparison distinguishes changing an answer from validating the change. It does not show that elicitation cannot help, or that a carefully specified population or persona prompt would fail. It shows that the available changes in question order and joint presentation do not establish a validity gain, even where they substantially change score dependence.

\subsection{Robustness and Measurement Resolution}
\label{sec:robustness}
Restricting to wave 2, weighting each record equally, or leaving out scenarios without excluding participants does not establish a positive sharing increment. The largest positive point estimate among these main-model sensitivity analyses is about 1.21\%; not every estimate is below one percent. The general pattern is small estimated gains with intervals that do not establish their sign, rather than a precisely identified zero effect.

A metadata check during manuscript preparation identified the 13 inconsistent participant identifiers described in Section~3.1. Excluding their records leaves 1,195 observations from 304 people across all 290 articles and 99 scenarios. Repeating the conditional sharing analysis gives fixed-fit increments of $-0.24$\% to $+0.67$\% and selected increments of $-1.63$\% to $+0.05$\%, with every interval crossing zero. This post-hoc sensitivity preserves the main assessment; it does not verify respondent identities or resolve the causes of the inconsistent entries.

In the reverse-outcome diagnostic, we predict human credibility from human sharing and model sharing, then add model credibility. Fixed-fit gains are positive in all eight versions, ranging from 1.77\% to 4.64\%, but their crossed-resampling intervals also cross zero. This is a descriptive asymmetry, not a successful positive control proving that the protocol can reliably distinguish one construct from the other.

Sparse human ratings particularly constrain article-mean interpretation. In a separate noise diagnostic on 289 eligible articles, the observed sharing error of a held-out mean baseline is 1.0835, while the estimated contribution of reader noise to the article means is 1.1161. The resulting corrected estimate is $-0.0326$, with an exploratory interval of $[-0.2863,0.2228]$. Negative squared error is not possible; the estimate shows that the between-article component is poorly resolved under that correction's assumptions. It should not be used as a denominator for a fraction-of-ceiling claim. The individual-record criterion avoids treating noisy article means as exact ground truth, but it does not create additional independent readers or resolve missing personal context.

\section{Discussion}
\subsection{What Does the Additional Question Provide?}
The sharing question adds numerical variation beyond observed credibility, and part of that variation recurs across formats. A stable increment for human sharing remains unestablished. These findings describe different properties of the same answers: neither output differentiation nor persistence supplies the missing external validation.

This decomposition makes the empirical claim more specific than saying that LLM answers are correlated or unreliable. A low correlation between two scores could reflect irrelevant noise; a high correlation could coexist with a small useful increment. Likewise, shared residual patterns across models could arise from either valid content sensitivity or common bias. For an evaluator built from multiple survey items, the relevant question is how each item changes the evidence available for its intended outcome, after accounting for the scores already collected.

The raw reversal remains informative as a warning about interpretation. It demonstrates how an intuitively aligned question can produce numerically worse predictions when its answer is used directly. Calibration and the single-human comparison explain why this observation cannot carry a broader claim that LLMs uniquely fail to understand sharing. The more durable lesson is that numerical calibration, score differentiation, and criterion validity require separate evidence.

The increment is conditional on the score pair obtained under a given format. Appending a question could also change the earlier credibility answer, so our score-inclusion comparison is not a randomized estimate of the causal effect of lengthening a questionnaire. The format analysis evaluates the resulting pairs separately. This distinction matters when designing a measurement battery: an additional field and an additional validated measurement are separate claims.

\subsection{Implications for Web Evaluation}
Researchers may reasonably want to evaluate misleading articles along several dimensions, including credibility and willingness to share. Our results support an explicit validation step before treating those outputs as distinct empirical measures. Define the additional outcome, compare against the measurements already available, and test the increment on held-out content. When human records are reused across articles, keep participant dependence visible in both fitting and uncertainty assessment. A richer questionnaire can produce richer-looking output without a demonstrated gain for the human criterion.

The appropriate comparison also depends on the intended use. Our conditional diagnostic is useful for studying whether the model score supplies information beyond observed human credibility. The model-only comparison is closer to a setting without human answers at prediction time. A material-screening application instead needs validation of the selected articles at its actual budget. Existing top-20\% screening results show positive descriptive lift in observed human sharing (Appendix~\ref{app:screening}), which can coexist with an unestablished increment over credibility. Absolute usefulness and incremental usefulness are different questions.

Fact-checking and platform interventions require validation of their own outcomes and costs. Our immediate implication concerns measurement: semantic correspondence and cross-prompt consistency describe the elicitation, while human criterion validation supports interpretations of its scores.

\subsection{Explanations Left Open}
Several accounts remain possible: the scores may expose different quantizations of a common evaluation; sharing may reflect article properties with weak additional human associations; its relevance may depend on personal interests, audiences, or social context absent from the prompt; or the sample and predictors may not resolve a small or nonlinear increment.

Finer response scales, repeated calls, controlled personal context, and denser independent human ratings would test different accounts. Current residuals do not identify a personality, intention, or causal pathway. The observed scale provides a measurement interface rather than direct access to internal computation.

\section{Limitations, Ethics, and Reproducibility}
The synthetic English corpus covers 99 selected scenarios and five topics, with generation explicitly targeting human judgments. Its content and convenience sample do not represent the full Web information environment. Uneven reader coverage and unverified demographic entries constrain generalization. The outcome is reported willingness under a fixed questionnaire order, not sharing in a natural feed.

One retained answer per model--article condition prevents estimating full generation variability. Six of eight versions share one family. Cross-format persistence is not a validated psychometric reliability coefficient. Humans complete a longer instrument, while models answer two items without profiles. These boundaries apply to the evaluated elicitation and need not hold for other protocols.

Seven-level scales are treated numerically, regression families are finite, and resampling conditions on fitted predictions. Prior data exploration and multiple comparisons preclude confirmatory significance claims. Small MSE increments do not bound information or establish equivalence; conditioning on human credibility leaves its measurement error and omitted personal factors unresolved.

The study received institutional ethics approval (H2025061; December 15, 2025). Participants gave informed consent, participated voluntarily, and received compensation through Prolific. Analysis records exclude platform identities. Synthetic falsehoods are research materials, not factual reporting. This audit uses frozen responses without new recruitment or model calls. Saved scores, protocols, fold predictions, and scripts permit reproduction without proprietary endpoints. Generative AI tools assisted manuscript preparation and analysis-code development; the authors remain responsible for the analyses and final text.

\section{Conclusion}
We examine what a sharing question adds to LLM credibility judgments of misinformation. The raw sharing disadvantage changes with calibration and is not a sufficient diagnosis of model information. Sharing answers contain variation beyond the observed credibility score, with some correspondence across elicitation formats and models. Yet the tested predictors do not establish an additional human-sharing benefit, including when format changes reduce coupling between the two answers. These findings distinguish a different answer, a persistent pattern across prompts, and a valid additional measurement. Each elicited outcome requires evidence for its intended interpretation.
\label{sec:mainend}

\clearpage
\bibliographystyle{ACM-Reference-Format}
\bibliography{references}

\appendix
\setcounter{dbltopnumber}{2}
\section{Elicitation and Data Scope}
\label{app:elicitation}
Main evaluator identifiers are \nolinkurl{anthropic/claude-sonnet-4.5}, \nolinkurl{google/gemini-3-pro-preview}, \nolinkurl{gpt-4.1-2025-04-14}, \nolinkurl{gpt-4o}, \nolinkurl{gpt-5-2025-08-07}, \nolinkurl{gpt-5.1-2025-11-13}, \nolinkurl{gpt-5.2-2025-12-11}, and \nolinkurl{gpt-5.4-2026-03-05}. No top-$p$, seed, completion-token limit, or reasoning configuration was explicitly set in the main calls. Requests used a 120-second timeout, with up to four attempts for transient or parsing failures. Scores are the retained successful pair, not averages over repeated draws.

The user message introduced an online reading survey and asked for a reader's first impression, without looking up the article. It stated that there were no right or wrong answers and asked the model to answer as in daily life. After the text, the focal items were:
\begin{quote}
\textbf{Credibility:} Overall, do the main claims in this text feel believable and realistic?\\
\textbf{Sharing:} If you saw this text in daily life, would you personally want to forward or share it with others?
\end{quote}
Credibility anchors were: 1, not at all believable; 2, mostly unbelievable; 3, somewhat unbelievable; 4, unsure/neutral; 5, somewhat believable; 6, mostly believable; and 7, very believable. Sharing anchors were: 1, not at all willing; 2, unlikely to share; 3, slightly unlikely; 4, unsure/neutral; 5, slightly likely; 6, likely to share; and 7, very willing. The output was a JSON object with integer fields \texttt{credibility} and \texttt{willingness\_to\_share}. The same item wording and anchors were used for humans.

The human attention check requested selecting 7. The complete instrument (Appendix~\ref{app:human-instrument}) also included understanding, perceived intent strength, a conditional intent-category question, self-reported belief change, and topic familiarity. Intent strength concerns the text's intent, not the respondent's sharing motivation; belief change is a retrospective report, not a measured pre--post difference. These additional responses are not covariates in the main diagnostic. The release provides cleaned focal responses, not a population sampling frame or complete recruitment exclusion ledger.

Archived allocation code documents sequential cycling in a February 2026 version and a March 25 revision prioritizing texts with fewer valid responses, rotating ties and avoiding already submitted texts for that identifier. Validity checks used age, country, display status, and attention checks, not credibility or sharing score magnitude. Deployment history and assignment logs are unavailable, so code dates do not establish which policy governed each wave; allocation cannot be described as independently randomized. Respondents supplied anonymous codes to link repeated submissions. Platform-ID fields are empty in the available local response export, preventing independent identity reconciliation.

For the metadata audit, country strings were lowercased and common spelling or naming variants combined (e.g., UK, England, and Scotland under United Kingdom). The demographic summary uses the first retained allocation per study identifier. The exclusion sensitivity removes the union of identifiers with more than one reported age or normalized country, without changing the original frozen data. It was added after the discrepancy was noticed during manuscript preparation and was not part of the earlier analysis protocol. The source records do not establish why these entries differ, how every article allocation was determined, or the full pre-cleaning recruitment denominator.

The supplementary format data contain credibility-first, sharing-first, credibility-only, and sharing-only calls. The two single-item calls form the separate condition. Analyses use paired common article sets; the Gemini supplementary model is Gemini 3.1 Pro Preview, distinct from the main Gemini 3 Pro Preview. All comparisons retain this distinction.

\section{Estimator and Resampling Details}
\label{app:estimators}
For a score $v\in[1,7]$, the category basis is $B_k(v)=\max\{0,1-|v-k|\}$, $k=1,\ldots,7$. At integer values it is a one-hot category representation; it linearly interpolates the exceptional averaged human response. An intercept and six basis terms identify unrestricted category means. The human conditioning response and the wave indicator are unpenalized. The fixed model adds raw numerical model scores using weighted least squares.

Flexible candidates have: (i) no model-score terms; (ii) linear terms; (iii) linear and squared terms, with the product of model scores when both are available and each score's interaction with the centered human conditioning response; or (iv) separate category bases for the model scores. The model terms are standardized using training-weighted means and standard deviations, with a unit fallback for a constant feature. Ridge penalties are $\lambda\in\{0,1,10,100\}$. Penalization is applied after projecting out the unpenalized covariates; $\lambda=0$ uses least squares. All fitted predictions are clipped to $[1,7]$ before computing losses.

Reconstruction uses the same single-score candidate families without human covariates. The baseline class for external validity contains the conditioning covariates and either no model score or the original score. The expanded class contains every baseline candidate and candidates with both scores. Five-fold inner losses select a candidate; ties after rounding to 12 decimal places follow a fixed order beginning with the no-model-score candidate. Outer folds are deterministic scenario-grouped splits. The individual-record analysis balances folds using the record grouping, whereas reconstruction operates on one row per article; both keep scenarios intact.

For each outer training set, inner validation holds out scenarios and excludes all inner-validation participants from inner training. Training participant weights use remaining training records; validation weights are inherited from participant weights in the outer training set. The final risk uses each participant's total number of evaluated records. This preserves the intended equal-person weighting while preventing response reuse across the training boundary.

The crossed bootstrap independently draws $P$ participant IDs and $K$ scenario IDs with replacement. If their multiplicities are $a_p$ and $b_{k(i)}$, the resampled risk uses weights $a_pb_{k(i)}w_{pi}$, normalized to sum to one. Gains use the same draw in both risks. Residual correlations and reconstruction intervals instead draw scenarios, include all their articles with multiplicity, and hold fitted quantities fixed. All analyses use 5,000 draws with a fixed seed. We do not claim full-pipeline bootstrap coverage or multiplicity-adjusted inference.

\section{Per-Model Results and Supplementary Comparisons}
\label{app:tables}
Table~\ref{tab:allmain} provides the exact main comparisons underlying Figure~\ref{fig:information}. Table~\ref{tab:aux} reports the model-only sharing increment and the reverse conditional diagnostic. A negative increment denotes higher held-out error for the expanded predictor. A positive point estimate alone does not establish a benefit.

\begin{table*}[t]
\caption{Main per-model results. Reconstruction uses nested selection; its RMSE is in score points. The last two columns report human-sharing MSE reductions from adding model sharing to human and model credibility and wave. Brackets are exploratory 95\% crossed-resampling intervals.}
\label{tab:allmain}
\centering
\begin{tabular}{lrrrr}
\toprule
Evaluator & Reconstruction (\%) & Residual RMSE & Fixed increment (\%) & Selected increment (\%) \\
\midrule
Claude 4.5 & 63.4 & 0.439 & -0.06 [-0.94, +0.86] & -0.05 [-0.15, +0.02] \\
GPT-4.1 & 54.3 & 0.819 & -0.18 [-1.48, +1.06] & -0.58 [-3.01, +1.50] \\
GPT-4o & 72.5 & 0.667 & +0.33 [-1.40, +2.01] & -0.60 [-2.49, +0.95] \\
GPT-5 & 48.5 & 0.727 & -0.25 [-1.07, +0.46] & +0.09 [-0.25, +0.61] \\
GPT-5.1 & 37.3 & 0.898 & +0.69 [-1.67, +2.71] & +0.44 [-1.76, +2.43] \\
GPT-5.2 & 32.7 & 0.640 & +0.30 [-1.14, +1.71] & +0.33 [-0.85, +1.45] \\
GPT-5.4 & 30.8 & 0.741 & +0.27 [-1.05, +1.68] & -0.26 [-1.69, +1.11] \\
Gemini 3 Pro & 66.4 & 0.640 & -0.02 [-1.31, +1.25] & -0.36 [-1.56, +0.37] \\
\bottomrule
\end{tabular}
\end{table*}

\begin{table*}[t]
\caption{Additional human-validity comparisons. Left: adding model sharing to model credibility and wave, without human credibility (selected fit). Right: adding model credibility to human sharing, model sharing, and wave to predict human credibility (fixed fit). Each column has its own baseline and target; differences between columns are not tests of construct superiority.}
\label{tab:aux}
\centering
\begin{tabular}{lrr}
\toprule
Evaluator & Model-only sharing increment (\%) & Reverse credibility increment (\%) \\
\midrule
Claude 4.5 & +0.03 [-0.49, +0.56] & +2.03 [-1.90, +6.70] \\
GPT-4.1 & -0.04 [-1.91, +1.66] & +2.82 [-2.44, +8.05] \\
GPT-4o & +0.23 [-0.26, +0.80] & +3.12 [-2.33, +8.71] \\
GPT-5 & +0.01 [-0.14, +0.17] & +2.28 [-2.54, +6.84] \\
GPT-5.1 & +0.20 [-0.57, +0.94] & +4.64 [-1.42, +10.35] \\
GPT-5.2 & +0.08 [-0.64, +0.77] & +4.64 [-1.03, +10.24] \\
GPT-5.4 & +0.26 [-0.37, +1.04] & +3.90 [-1.55, +9.12] \\
Gemini 3 Pro & +0.25 [-0.56, +1.06] & +1.77 [-1.90, +5.18] \\
\bottomrule
\end{tabular}
\end{table*}

\subsection{Sensitivity Analyses}
\label{app:sensitivity}
We repeat the main conditional sharing comparison after restricting to wave 2, weighting records equally, and omitting participant exclusion while retaining scenario holdout. In every case, both fixed and selected fits evaluate all eight main versions, and every individual interval crosses zero. The largest positive point estimate is about 1.21\%; sensitivity analyses do not justify a universal numerical bound below one percent. Inner selection still occurs entirely within outer training data.

The reverse-outcome diagnostic conditions on human sharing and model sharing before adding model credibility to predict human credibility. Its all-positive fixed-fit point estimates should not be confused with a statistically established positive control. The target, baseline information, and measurement precision differ from the main comparison. Adding model sharing to a model-credibility predictor of human credibility is a further auxiliary check, also insufficient to establish a consistently positive increment. These results are not used to choose the primary outcome after observing its significance.

\subsection{Article-Mean Noise and Human Reference}
For an article with $n_i\geq2$ readers and sample sharing variance $s_i^2$, subtracting $s_i^2/n_i$ from $(\widehat f_i-\bar S_i^h)^2$ estimates latent-mean squared error under conditionally independent, identically distributed reader responses and predictor independence from test noise. We apply this diagnostic to 289 articles and scenario-and-participant-held-out predictions. The main text reports its imprecise, negative estimate for the mean baseline. Shared readers and heterogeneous allocation can complicate the assumptions, so this estimate is not an identified predictive ceiling.

For the single-human reference, each reader's raw $C$ or $S$ predicts the mean sharing response of the other readers of that article. We average squared errors over the possible held-out-reader choices and then equally over eligible articles. Thus the prediction and reference do not reuse the same response. The raw MSEs, 4.772 and 5.833, quantify this particular task. They do not imply that human credibility is generally a better measure of sharing intentions, or that the human and model prediction tasks are identical.

\section{Screening as a Different Use of Scores}
\label{app:screening}
An existing exploratory analysis ranks articles by a frozen model score and selects a fixed fraction $q$. Boundary ties receive fractional inclusion weights: $w_i=1$ above the cutoff, zero below it, and $r/t$ for each of $t$ tied articles when $r$ places remain. Therefore $\sum_iw_i=qN$, even for a noninteger expected budget. Lift is
\begin{equation}
L_q=\frac{\sum_iw_i\bar S_i^h}{qN}-\frac{1}{N}\sum_i\bar S_i^h.
\end{equation}
This is expected lift under uniform random breaking of boundary ties. Human labels do not choose the articles or resolve ties, and positive affine score transformations do not alter selection.

At a 20\% budget, sharing-score lift ranges from 0.242 to 0.365 points over expected random selection, and credibility-score lift from 0.263 to 0.428. Every paired $C$--$S$ lift interval crosses zero. The original intervals resample scenarios and rerank the pool; they do not fully address shared participants. These results describe selection within the current materials and do not validate recovery of the truly highest-sharing articles or benefits from intervention. They are compatible with the main finding: a score can associate with a useful outcome while supplying little established increment over another score already available.

\section{Human Survey Instrument}
\label{app:human-instrument}
The archived instrument contains the following pre-text instructions and questions, in the order shown. Response labels are reproduced verbatim; numbering and typography are normalized. Before the task, the consent page stated: ``Some texts may contain inaccurate information; others may be accurate.'' The displayed 15--20 minute duration was an estimate, not an observed completion-time statistic.

\paragraph{Reader instructions.}
``Imagine you come across the following text during everyday online reading. Please read it naturally (no need to read `extra carefully'), then answer based on your first impression. Please do not look it up online; answer only based on what is shown here.'' How to answer: ``Answer based on your impression as a reader.'' ``Do not try to guess the researchers' intent.''

\paragraph{1. Understanding.}
``Overall, how well do you think you understand the text you have just read?'' Anchors: 1 = Not at all; 2 = Barely; 3 = Some parts; 4 = Unsure / Neutral; 5 = Mostly; 6 = Very well; 7 = Completely.

\paragraph{2. Credibility.}
``Overall, do the main claims in this text feel believable and realistic?'' Anchors: 1 = Not at all believable; 2 = Mostly unbelievable; 3 = Somewhat unbelievable; 4 = Unsure / Neutral; 5 = Somewhat believable; 6 = Mostly believable; 7 = Very believable.

\paragraph{Attention check.}
``To confirm you are paying attention, please select 7 for this question.''

\paragraph{3. Willingness to share.}
``If you saw this text in daily life, would you personally want to forward or share it with others?'' Anchors: 1 = Not at all willing; 2 = Unlikely to share; 3 = Slightly unlikely; 4 = Unsure / Neutral; 5 = Slightly likely; 6 = Likely to share; 7 = Very willing.

\paragraph{4. Intent strength.}
``Do you think this text is written with a strong intent?'' Anchors: 1 = No intent at all; 2 = Very weak intent; 3 = Weak intent; 4 = Unsure / Neutral; 5 = Somewhat strong intent; 6 = Strong intent; 7 = Very strong intent.

\paragraph{4.2. Main intent categorization.}
Shown when intent strength is at least 2: ``From your perspective, which option best describes the text's main intent? These options describe influence attempts. If none fits, choose `Unclear / mixed'.'' Options:
\begin{itemize}
\item \texttt{opinion\_shaping}: Trying to steer opinions about a topic or group using claims that feel clearly misleading, exaggerated, or detached from reality
\item \texttt{attention\_harvesting}: Trying to catch an audience's attention
\item \texttt{unclear}: Unclear / mixed / no clear intent to influence
\end{itemize}

\paragraph{5. Belief change.}
``To what extent did this text change your prior understanding or beliefs about the topic?'' Anchors: 1 = Not at all; 2 = Very little; 3 = A little; 4 = Unsure / Neutral; 5 = Somewhat; 6 = A lot; 7 = A great deal.

\paragraph{6. Topic familiarity.}
``Overall, how familiar are you with the topic of the text you just read?'' Anchors: 1 = Not familiar at all; 2 = Slightly familiar; 3 = Somewhat unfamiliar; 4 = Unsure / Neutral; 5 = Somewhat familiar; 6 = Quite familiar; 7 = Very familiar.

\end{document}